# Autonomous Navigation by Robust Scan Matching Technique

D. Banerji[1], R. Ray[2], J. Basu[3], I. Basak[4]

[1,2]Scientist, CSIR-Central Mechanical Engineering Research Institute, Durgapur, India
{[1]dbanerji, [2]ranjitray}@ cmeri.res.in

[3]Professor, Mechanical Engineering Dept., Birbhum Institute of Technology, India
[3]b.jhankar@gmail.com

[4]Professor, Mechanical Engineering Dept., National Institute of Technology, Durgapur, India
[4]basak_indrajit@yahoo.com

*Abstract*—For effective autonomous navigation, estimation of the pose of the robot is essential at every sampling time. For computing an accurate estimation, odometric error needs to be reduced with the help of data from external sensor. In this work, a technique has been developed for accurate pose estimation of mobile robot by using Laser Range data. The technique is robust to noisy data, which may contain considerable amount of outliers. A grey image is formed from laser range data and the key points from this image are extracted by Harris corner detector. The matching of the key points from consecutive data sets have been done while outliers have been rejected by RANSAC method. Robot state is measured by the correspondence between the two sets of keypoints. Finally, optimal robot state is estimated by Extended Kalman Filter.

The technique has been applied to an operational robot in the laboratory environment to show the robustness of the technique in presence of noisy sensor data. The performance of this new technique has been compared with that of conventional ICP method. Through this method, effective and accurate navigation has been achieved even in presence of substantial noise in the sensor data at the cost of a small amount of additional computational complexity.

## I. INTRODUCTION

For effective navigation of a mobile robot, pose estimation of the robot in the environment is an essential task. The purpose of the pose estimation is to keep track of the position and heading of mobile robot with respect to a global reference frame. By using the encoder data, pose of the robot can be computed, but due to various systematic and unsystematic odometric errors like wheel slippage etc, error is accumulated without bound. Practically, within a short period, the pose error becomes so high that it cannot be used for a purposeful navigation. To overcome this problem, the robots pose needs to be computed with some suitable external reference at a specific sampling time interval. In this context, the relevant external references may be the natural landmarks or a set of distinguishable features present in the environment, which do not change position with time. For acquisition of these relevant data from the environment, various types of external sensors are used by mobile robots. In this work Laser range data has been used for the perception of the environment.

Laser data have been used in various ways in different techniques for navigation purpose by scientists and researchers. The central issue of all these techniques is the proper correspondence of the data sets in the consecutive time steps. In other words, accurate data association is one of most important issue in this area. The problem becomes more critical when the data sets contain substantial amount of noise or outliers. One strength of Laser scan technique is that the entire sensor scan data may be considered as a feature, no information is lost and any arbitrary shape in the environment can be represented without aggregating all measurements within a given region into a single value [1]-[4]. But this involves processing of large amount of sensor data which make the computational and storing process more complex. The said complexity can be reduced by introducing occupancy grid concept [5-7]. Here data association which is the most critical part of simultaneous localization and view-based metric map building can be performed by various methods. Scan matching is one of the important methods which are dealt with in the present work.

In the scan matching approach, the full batch of data obtained from local sensors is associated with the global map data in terms of correlations between the two sets. However, the scan matching technique is applied in a variety of ways for mobile robot navigation. Puttkamer E. et al. [8]-[9] explored a cross-correlation based scan matching technique from the derivatives of range-finder to determine the position and orientation without any geometric relationship with any feature. Tomono M. [2] proposed a geometric hashing scheme based global localization technique applicable to an environment having many curved features. Lu F. and Milios E. [1] developed a local scan matching approach by LRF data considering the relative pose of the robot. Albert Diosi and Lindsay Kleeman [3] proposed a Scan Matching approach in the polar coordinate system of a laser scanner, thereby eliminating expensive search for corresponding points in other scan match approaches. Censi A. et al. [10] suggested a feature data matching approach in Hough domain. Another approach is the correlation based image intensity based occupancy grid map building [11] where the matching problem can be solved by the establishment of the correspondence between two images transformed from the LRF data using occupancy grid concept. However, the most popular method is the Iterative Closest Point (ICP) algorithm [12-13] based on an iterative process to compute the correspondences between the scans, and then compute the sensor displacement by minimizing the distance error. Besl and McKay [12] described a general-purpose iterative closest point (ICP) algorithm for shape registration based on closest point rule and also proved that the ICP algorithm always converges monotonically to a local minimum with respect to the least-squares distance function. Moreover, regardless of the type of the model, the convergence speed of the algorithm is always very slow when the distance function approaches a local minimum. To accelerate the ICP algorithm, Besl and McKay used a line search method to heuristically determine the transformation variables based on their values in two or three recent iterations. Although this improves the convergence speed near a local minimum, the problem of obtaining a poor solution for the rotation component still exists. It has been observed that, if the rotation is small, the ICP algorithm is good at solving the translation. Again, in ICP method which is based on least square minimization technique, best matching result depends on the fit of the model data with the data. If the data sets having a lot of outliers/noise, the final matching result will be definitely influenced and will produce an erroneous matching. Hence the match technique is less robust. In [3] [13], some pre-processing technique was adopted to reduce the outliers. But, still rejections of all the outliers/noise from the data are very tedious job. So, rather than believing on the fit of all the elements of model with data, it's better to match by fitting some key elements from the data. Here, key elements may be defined as corner points or interest points. In this paper, an attempt is made to accurately measure the global pose of a mobile robot by a new technique of laser scan matching and to use this measurement for optimal pose estimation.

The deliberations of the paper are organized as follows. In section II, the kinematic model for pose prediction has been stated. Then, the process of image formation from LRF data, key point extraction and matching has been discussed and finally robot pose has been found by RANSAC method. In section III, the odometric information & measurement information has been fused by Extended Kalman filter for optimal estimation of pose. Subsequently, the testing & experimental results are elaborated in section IV. Finally, conclusion has been drawn in Section V. The scheme of the proposed method is shown in Fig. 1.

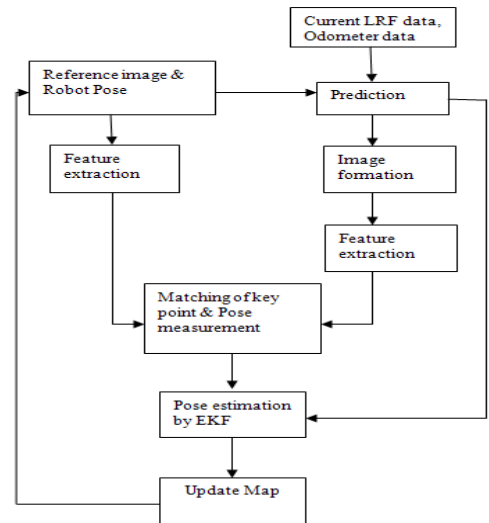

Fig. 1. Schematic diagram of the proposed method

II. MEASUREMENT OF ROBOT POSE

For a differential drive system the pose of the robot is predicted by the following kinematic equations,

$$x_{i+1} = x_i + \Delta S \cos(\theta_{i+1}) \quad (1)$$
$$y_{i+1} = y_i + \Delta S \sin(\theta_{i+1}) \quad (2)$$
$$\theta_{i+1} = \theta_i + \Delta\theta \quad (3)$$

where, $\{x_i, y_i, \theta_i\}^T$ is the pose at $i^{th}$ instant and $\{x_{i+1}, y_{i+1}, \theta_{i+1}\}^T$ is the predicted pose at $(i+1)^{th}$ instant after giving the inputs to the driving wheels. The predicted pose vector $\{x_{i+1}, y_{i+1}, \theta_{i+1}\}^T$ will be denoted by $X'_{i+1}$ in section IV.

$$\Delta S = \frac{\pi D(N_L + N_R)}{2nr_e} \quad (4)$$

$$\Delta\theta = \frac{\pi D(N_L - N_R)}{2nr_e B} \quad (5)$$

where, D= wheel diameter, B= vehicle width, n = gear ratio, $r_e$ = encoder resolution (pulse per revolution)

And $N_L, N_R$ are the inputs to the driving wheels and these are obtained from encoder data of the left and right wheels of the robotic vehicle at that sampling time.

Now, due to various types of systematic and unsystematic odometry errors like wheel slippage, unequal wheel diameters etc the pose computed by equations (1),(2) & (3) is erroneous. Next, the robot pose is measured with LRF data by a new technique in the following part of the section.

The range data acquired from the LRF is transformed into a local occupancy grid map and then it is converted into an image. It is the fact that all the identical feature points on a particular object may not be usually traced by the LRF in successive scanning. Identical feature points may be within very close proximity, but not exactly in the same position for two successive scans. Moreover, there may be occlusions. These anomalies, in fact, create the outliers. Presence of outliers can be drastically reduced if the features are considered as an area object instead of a point. This is a novel method for reducing outliers. Two sub-images are constructed from the current and previous instant (reference) of LRF data and then key points are extracted utilizing 'Harris corner point' detection method [14] as described later on. Based on the two set of key points, current image is aligned with the reference image using Random Sample Consensus (RANSAC) method [15].

*A. Construction of Image from LRF data*

From the real time navigation point of view, occupancy grid framework is more robust and unified approach compare to any other framework. Also grid concept reduces the possibility of outliers in an image. In this work, occupancy grid framework is adopted, but in different form. Here, the tessellated space is transformed to a gray image considering each cell (a × a) of the tessellated space to be a pixel and the pixel intensities are determined based on whether the cell is occupied or unoccupied. 2D Laser data $\{d_i, \theta_i\}_{i=1}^n$ are converted from the polar to the Cartesian form as written in (6). Here the assumption is that LRF is mounted at the geometric centre of the robot.

$$\begin{aligned} x_i &= x_c + \delta x + d_i \cos(\theta_i + \varphi) \\ y_i &= y_c + \delta y + d_i \sin(\theta_i + \varphi) \end{aligned} \quad (6)$$

where ($x_c$, $y_c$) is the previous robot position derived from the reference image map and (•x, •y, •) is the current increment as shown in Fig. 2. ($x_i$, $y_i$) is the Cartesian global position of the point object $P_i$ which is acquired by the LRF as $(d_i, \theta_i)$ in local polar coordinate.

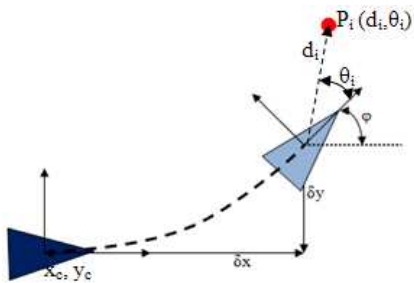

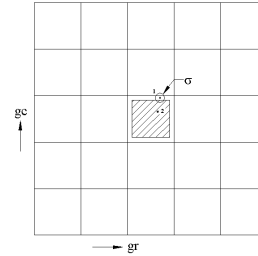
Fig. 2. Encoder based robot pose at $t^{th}$ & $(t+\bullet t)^{th}$ instant

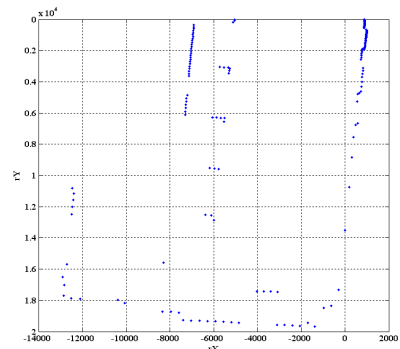
Fig. 3. Uncertainty in grid occupancy

If the LRF data is taken as true then occupied cells can be calculated considering cell size of (a×a) as

$$\begin{aligned} gr_i &= (\text{int})(x_i - a/2)/a + 1 \\ gc_i &= (\text{int})(y_i - a/2)/a + 1 \end{aligned} \quad (7)$$

having robot (LRF) position at cell {0, 0}. LRF data is highly accurate compared to sonar and other acoustic sensors, but still have some noise with standard deviation •.

Therefore, during conversion from cartesian point obtained from LRF data to cell, there is a probability to occupy more than one cell. As shown in Fig. 3, the point marked 1 is so placed that the uncertainty circle is overlapped on two consecutive cells.

The point may lie on any one of the two cells, whereas the point marked 2 is well inside the cell. Here, a simple heuristic rule is formulated as follows:

if, [ $\{(gr_i - 1)a - a/2 + \sigma\} < x_i \leq \{(gr_i - 1)a + a/2 - \sigma$
&& $\{(gc_i - 1)a - a/2 + \sigma\} < y_i \leq \{(gc_i - 1)a + a/2 - \sigma$ ]
  $I(gr_i, gc_i) = 255$
else
  $I(gr_i, gc_i) = 0$   ;
End

Hence, an image is being formed by considering each cell of size (a × a) equivalent to one pixel having a value either 0 if empty or 255 if occupied, as shown in Fig. 4.

Fig. 4. Plot of 2D Laser data into a tessellated space

In the figure, white cells represent the regions which are occupied by the features and black region shows the empty regions in the environment.

The number of rows and columns of the image are computed as follows,

$$rowT = \max\{gr_i\} - \min\{gr_i\} + 1$$
$$colT = \max\{gc_i\} - \min\{gc_i\} + 1 \quad (8)$$

and the center of robot is given as

$$Robot\_CenX = -\min\{gr_i\}$$
$$Robot\_CenY = -\min\{gc_i\} \quad (9)$$

### B. Key point extraction

Image is formed by transforming cartesian points derived from LRF data to image pixel as described earlier. In this transformation, each pixel represents a constant area in XY plane. Size of any object in the image containing number of point features will remain constant though their orientation may change because of different view angle. Hence, there is no scale variation among the features within the images developed in different time instant. Only the orientation has changed depending on the robot's pose. Therefore, conventional Harris corner detector which is invariant to rotation and computationally very fast, is sufficient to evaluate the stable key points. In this work, Harris corner point detector technique [14] is adopted.

### C. Data association by RANSAC method

Data association is one of the most difficult area of map building. Finding the correct correspondences between the previously stored key points and those extracted from current sensor data is a complicated task. Especially, it becomes more challenging in the presence of huge amount of noise to find the correct matches with least error in statistical sense. However, it is true that any image is always having some type of noises and consequently, the key points extracted from the image are associated with positional uncertainty. Though, majority of the points are having Gaussian error, but there is a fair chance that a number of points are extremely deviated and logically these points should be considered as outliers. Otherwise, the accuracy of the result would be critically degraded. By applying RANdom SAmple Consensus these outliers are removed and the pose of the robot is computed indirectly with least error.

Here, localization problem has been formulated as a hypothesis testing problem, where multiple pose hypotheses are considered and only the pose which can match the maximum number of features or key points in the current sensor data is accepted as the best probable candidate for the pose.

The idea of RANSAC is to generate a set of model hypotheses by selecting randomly subsets of the extracted key points containing the minimum number of data points sufficient to define a model. For each hypothesis a set of data points which fits the hypothesized model within a suitable tolerance is determined, called the consensus set. The hypothesis corresponding to the maximum consensus set is considered to be the most probable one. Here, minimum one pair of key points at a time from the image is used to generate model hypothesis. The method has been applied as follows.

*1) Initial Key Point matching:* For the initial correspondence, the key points on both the current image $\{X_c, Y_c\}$ and reference image $\{X_r, Y_r\}$ are projected to the global image map and then compared. Here, association of key points between current and reference image largely depend on the process noise. With lower process noise, two sets of data to be matched are within close vicinity and the corresponding association problem becomes simple. But in real situation, especially when the vehicle takes turn, large process noise is generated due to skidding. The odometer data is utilized for computing the search space for the correct key point. In this work, tentative key points are associated as follows.

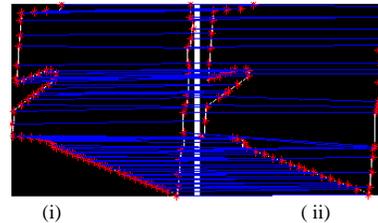

(i)　　　　　　　　( ii)
Fig. 5. Tentative keypoints matching between i) Reference & ii) Current Images

For a reference key point, nearest neighbourhood key point in current image is searched out using the euclidean distance as follows:

$$d_{min} = \sqrt{(X_r - X_c)^2 + (Y_r - Y_c)^2} \quad (10)$$

Now searching for all key points in current image within the range $x \in [X_{\bar{c}} - w, X_{\bar{c}} + w]$ and $y \in [Y_{\bar{c}} - w, Y_{\bar{c}} + w]$ around the nearest neighbourhood key point $(X_{\bar{c}}, Y_{\bar{c}})$ in current image is done, where w depends on the process noise. In this method of tentative matching, sometimes, more than one correspondence may be established for a reference key point as shown Fig. 5. The fake correspondences are rejected in the subsequent steps and only the correct match is retained.

*2) Computation of the tentative robot poses:* Two tentative matches are selected randomly from the tentative matching list and then evaluated the alignment parameters $(X_{co}, Y_{co}, \phi_{co})$ as follows:

From the first pair,

$$\begin{bmatrix} X_{co} \\ Y_{co} \end{bmatrix} = \begin{bmatrix} X_i \\ Y_i \end{bmatrix} - \begin{bmatrix} \cos\phi_{co} & \sin\phi_{co} \\ -\sin\phi_{co} & \cos\phi_{co} \end{bmatrix} \begin{bmatrix} X_j \\ Y_j \end{bmatrix} \quad (11)$$

$$\begin{bmatrix} X_{co} \\ Y_{co} \end{bmatrix} = \begin{bmatrix} X_i' \\ Y_i' \end{bmatrix} - \begin{bmatrix} \cos\phi_{co} & \sin\phi_{co} \\ -\sin\phi_{co} & \cos\phi_{co} \end{bmatrix} \begin{bmatrix} X_j' \\ Y_j' \end{bmatrix} \quad (12)$$

where, $P_i(X_i, Y_i)$ and $P_i'(X_i', Y_i')$ are two randomly selected key point positions on reference image and $P_j(X_j, Y_j)$ and $P_j'(X_j', Y_j')$ are the corresponding pairs on current image. By equating two matches, the following equations are derived:

$$\begin{bmatrix} C \\ D \end{bmatrix} = \begin{bmatrix} \cos\phi_{co} & \sin\phi_{co} \\ -\sin\phi_{co} & \cos\phi_{co} \end{bmatrix} \begin{bmatrix} A \\ B \end{bmatrix} \quad (13)$$

where, $A = X_j - X_j'$, $B = Y_j - Y_j'$, $C = X_i - X_i'$, $D = Y_i - Y_i'$

Now if the tentative matches are correct, then the relative positions of Pi, Pi' and Pj, Pj' are invariant and hence $A^2 + B^2 \approx C^2 + D^2$. This method facilitates the early elimination of wrong matches. From (13), $\phi_{co}$ is obtained as shown in (14).

$$\phi_{co} = \tan^{-1}\left(\frac{BC - AD}{AC + BD}\right) \quad (14)$$

By substituting this value in (12), $(X_{co}, Y_{co})$ are obtained.

*3) Computation of support:* Now, it is checked how many tentative matches support the tentative robot pose $(X_{co}, Y_{co}, \theta_{co})$.

First the current key point positions $(X_{jp}, Y_{jp})$ in the matching list are computed for each match k by (15).

$$\begin{bmatrix} X_{jp} \\ Y_{jp} \end{bmatrix} = \begin{bmatrix} \cos\phi_{co} & -\sin\phi_{co} \\ \sin\phi_{co} & \cos\phi_{co} \end{bmatrix} \begin{bmatrix} X_k - X_{co} \\ Y_k - Y_{co} \end{bmatrix} \quad (15)$$

Then modified current key point positions are compared with the tentative matching pair of in the reference key point positions. If the deviation is within the limit, then it is considered as a supporting match.

*4) Hypothesis with Most support:* Steps from 1 to 3 are repeated n times. The alignment parameters $(X_{co}, Y_{co}, \phi_{co})$ with most support, is the selected hypothesis and it is considered as the best measurement of the robot pose. It is denoted by $X_{i+1}$ in the next section, where optimal pose is estimated.

The required number of iterations (n) is calculated from the probability of a good matching • for RANSAC as

$$\delta = 1 - (1 - (1-\varepsilon)^j)^n \quad (16)$$

where, • is the ratio of false matches to total matches, j is the sample size.

### III. ROBOT STATE ESTIMATION

In section III, we have indirectly measured the robot state from Laser data. Similar to the odometer error, the measurement obtained from LRF data is also associated with some Gaussian error, which is called measurement error. By using Extended Kalman Filter, the odometer data & observed data from Laser scan, is fused to compute the optimal estimate of the robot pose. It is done iteratively by prediction and correction of the robot pose. Here, the robot state is predicted by encoder data (equation 1-3) and the state covariance is also predicted as

$$P_{i+1}' = J_v P_i J_v^T + J_u Q J_u^T \quad (17)$$

where, $P_i$ is the state covariance at $i^{th}$ instant and $P_{i+1}'$ is the predicted state covariance at the $(i+1)^{th}$ instant respectively. $J_v$ and $J_u$ are the Jacobians of the nonlinear state transition function with respect to vehicle state and process noise respectively. Q is the process noise covariance.

Innovation is computed as
$$v_{i+1} = X_{i+1}' - X_{i+1} \quad (18)$$

where, $X_{i+1}'$ is predicted pose from equation (1) and $X_{i+1}$ is the observed pose from equations (12) - (14).

Now, Innovation covariance,
$$S_{i+1} = HP_{i+1}'H^T + R \quad (19)$$

where, $H$ is Jacobian of observation matrix. The measurement covariance $R$ is the residual error of the robot pose computed by the method [3] and it is ($(1p, 1p, 0.25^0)$), where p is pixel i.e., 50mm.

Kalman_gain, $W_{i+1} = P_{i+1}'H^T S_{i+1} \quad (20)$

Hence, the update of the robot pose and the associated covariance are computed by equation (21) & (22) respectively.

$$X_{i+1} = X_{i+1}' + W_{i+1}v_{i+1} \quad (21)$$
$$P_{i+1} = P_{i+1}' - W_{i+1}S_{i+1}W_{i+1}^T \quad (22)$$

After computing the updated robot pose, modified current sub-image is superimposed on the previous reference image by co-positioning the robot position at both the current and the previous reference images. For superimposing the current image on the global image, the reference image is zero padded if the local image falls beyond the boundary of the reference image at a particular robot position. This updated global image is now stored as the reference image for the next iteration.

## IV. EXPERIMENTS AND RESULTS

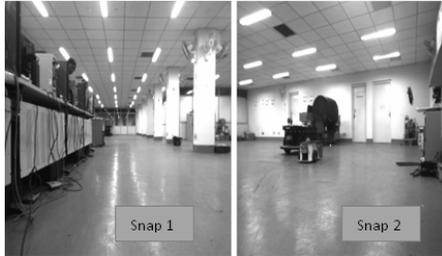

Fig. 6. Real environment for conducting the test at robotics laboratory

The performance of the proposed technique has been tested in the laboratory by conducting experiments with operational mobile robot. The performance has been compared with the conventional ICP method [16]. Real world data were captured by a Pioneer P3-DX robot with an on-board SICK LRF in the laboratory as shown in Fig. 6. The robot was moved through the working environment in a predefined path such that every possible corner of the environment could be explored. Over 3300 set of LRF and pose data were captured at an interval of 200ms with average speed of 200 mm/s.

The global map based on raw odometer and laser data is plotted in Fig. 7 which shows that it is associated with huge process noise.

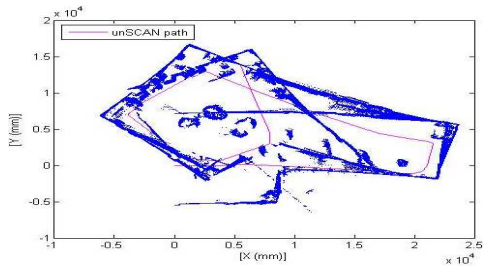

Fig. 7. Map generated by the data of odometer and LRF

All sample data sets were converted to image considering grid size 50mmX50mm and matched using ICP [16] and proposed method respectively without any rejection of outliers. Residual errors are computed with the help of the technique used in [3] taking unit weight factor for both the cases. Scan matching using ICP method produces larger errors when vehicle take sharp turn as shown in Fig. 8(i). ICP method has a natural tendency to converge towards local minima. In fact, it's a drawback of ICP method [1]. In the proposed method, few keypoints may be outliers and also some tentative matching may be erroneous. But, in due course, all the fake keypoints and miss-matching are eliminated in the form of outliers while seeking for maximum support in RANSAC method and finally converges to actual matching point as shown in Fig. 8(ii).

Residual errors are represented in the form of pixel. Here one pixel error is equivalent to maximum 50mm error. Residual errors are as per with different golden approach [3][13].

For testing the robustness of the proposed method random noise is gradually injected to the current LRF data set. It has been observed that up to 43% of noisy data out of 181 data in each scanning, the proposed method gives almost same result and after that residual errors are unpredictable.

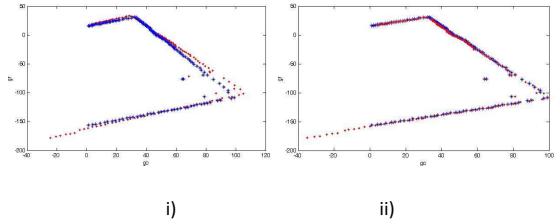

i)          ii)

Fig. 8. Match at the instant where (•x, •y, ••) = (43mm, 27mm, 14.63°) by i) ICP (30 iteration), ii) proposed method considering •=0.9, •=0.7, j=2, m=37, matches = 23. Here red dotted points depict the reference feature points whereas blue line for current. Residual error (•x, •y, ••) = (1p, 1p, 0.25°), where p is pixel.

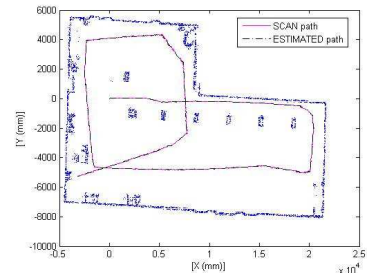

Fig. 9. Generated global image map

The trajectory of robot and the final global map is generated and plotted in Fig. 9. Here, the parameters are taken as 50mm x 50mm grid cell, • = 5 mm.

## V. CONCLUSION

The paper presented a new technique of autonomous localization by robust scan matching of LRF data. LRF data has been converted into a grey image and key points of the image have been used as landmarks for localization. By this process, the volume of data to be processed has been reduced considerably. By using RANSAC, the outlier data has been rejected and higher accuracy and robustness have been achieved. Experimental results has been compared with that of conventional ICP method to show the strength & usefulness of the proposed method.